\documentclass[acmlarge]{acmart}
\AtBeginDocument{%
  \providecommand\BibTeX{{%
    \normalfont B\kern-0.5em{\scshape i\kern-0.25em b}\kern-0.8em\TeX}}}

\usepackage{color}
\usepackage{multirow}
\usepackage{subfigure}
\usepackage{pifont}
\usepackage{svg}
\usepackage{CJKutf8}

\setcopyright{acmlicensed}
\acmJournal{Tallip}
\begin{document}

% \begin{CJK*}{UTF8}{gbsn}
\title{A Simple Yet Effective Corpus Construction Framework for Indonesian Grammatical Error Correction}

\author{Nankai Lin}
\email{neakail@outlook.com}
\affiliation{%
  \institution{School of Computer Science and Technology, Guangdong University of Technology}
  \city{Guangzhou}
  \state{Guangdong}
  \country{China}
  \postcode{510000}
}

\author{Meiyu Zeng}
\affiliation{%
  \institution{School of Computer Science and Technology, Guangdong University of Technology}
  \city{Guangzhou}
  \state{Guangdong}
  \country{China}
  \postcode{510000}
}

\author{Wentao Huang}
\affiliation{%
  \institution{School of Computer Science and Technology, Guangdong University of Technology}
  \city{Guangzhou}
  \state{Guangdong}
  \country{China}
  \postcode{510000}
}

\author{Shengyi Jiang}
\affiliation{%
  \institution{School of Information Technology and Engineering, Guangzhou College of Commerce}
  \city{Guangzhou}
  \state{Guangdong}
  \country{China}
  \postcode{510000}
}
\email{200511402@oamail.gdufs.edu.cn}

\author{Lixian Xiao}
\authornotemark[1]
% \authornote{Lixian Xiao and Shengyi Jiang are co-corresponding authors.}
\affiliation{%
  \institution{Faculty of Asian Languages and Cultures, Guangdong University of Foreign Studies}
  \city{Guangzhou}
  \state{Guangdong}
  \country{China}
  \postcode{510000}
}
\email{173829137@qq.com}

\author{Aimin Yang}
\affiliation{%
  \institution{School of Computer Science and Technology, Guangdong University of Technology}
  \city{Guangzhou}
  \state{Guangdong}
  \country{China}
  \postcode{510000}
}

\renewcommand{\shortauthors}{Lin et al.}

\begin{abstract}
  Currently, the majority of research in grammatical error correction (GEC) is concentrated on universal languages, such as English and Chinese. Many low-resource languages lack accessible evaluation corpora. How to efficiently construct high-quality evaluation corpora for GEC in low-resource languages has become a significant challenge. To fill these gaps, in this paper, we present a framework for constructing GEC corpora. Specifically, we focus on Indonesian as our research language and construct an evaluation corpus for Indonesian GEC using the proposed framework, addressing the limitations of existing evaluation corpora in Indonesian. Furthermore, we investigate the feasibility of utilizing existing large language models (LLMs), such as {GPT-3.5-Turbo} and GPT-4, to streamline corpus annotation efforts in GEC tasks. The results demonstrate significant potential for enhancing the performance of {LLMs} in low-resource language settings. {Our code and corpus can be obtained from https://github.com/GKLMIP/GEC-Construction-Framework.}
\end{abstract}





\maketitle

\section{Introduction}

Grammar consists of precise rules that formalize the relationships among multiple words \cite{baviskar2019comparative}. Grammatical errors diminish the readability of texts and adversely impact the reader's experience. So grammatical error correction (GEC) has become a significant research orientation within the field of natural language processing (NLP).
Numerous NLP researchers have primarily concentrated on GEC in universal languages like English and Chinese. Nevertheless, there is little research {especially} focused on Indonesian, which is categorized as a low-resource language belonging to the Malay-Polynesian branch of the Austronesian language family. Despite having a population of over 270 million speakers, Indonesian still has severely limited linguistic resources, which poses challenges for NLP research. And the limitation of resources adversely impacts the progress of Indonesian NLP technology and GEC research in Indonesian.
%\footnote{https://www.bps.go.id/pressrelease/2021/01/21/1854/hasil-sensus-penduduk-2020.html\#:~:text=Hasil\%20Sensus\%20Penduduk\%20\%28SP2020\%29\%20pada\%20September\%202020\%20mencatat,kepadatan\%20penduduk\%20Indonesia\%20sebanyak\%20141\%20jiwa\%20per\%20km2.} 

Currently, limited attention has been concentrated on GEC tasks for Indonesian language processing. \citet{10.1145/3440993} proposed an Indonesian GEC framework, which checked and corrected a word of one certain POS in the sentence by choosing recommended words from the designed POS confusion sets. They confirmed the effectiveness of the framework in handling low-resource GEC {tasks}. However, it is worth noting that \citet{10.1145/3440993} regarded the GEC task as a multi-classification problem, which imposed certain limitations and rendered it unsuitable for evaluating generative-based GEC models. \citet{app122010380} introduced a universal GEC model based on the Transformer architecture. The model was not only adaptable for grammatical correction in Indonesian but also could be applied to other low-resource languages. Nevertheless, the evaluation corpus they used consisted of synthetic data without manual annotation. Consequently, it becomes challenging to measure the model’s accuracy reliably, leading to potential instability in performance evaluation. Moreover, there are differences in the distribution between synthetic data and real-world data, which means that errors in synthetic data may not necessarily be reflected in real-world data.

Besides, \citet{10.1162/coli_a_00478} emphasized the important role of corpus construction for GEC tasks and also highlighted it as a critical basis within the GEC tasks. In the case of low-resource languages, the absence of a language learner platform for acquiring grammatical error correction materials poses a challenge. Consequently, the primary challenge of {the} GEC task lies in constructing {an} annotated corpus that {conforms} to the distribution of real-world data.

In this paper, we introduce a corpus construction framework and develop an Indonesian GEC evaluation corpus. Specifically, our approach involves training an Indonesian GEC model using faulty synthetic data, {and} subsequently applying the trained model to correct errors in real-world data. For the sentences modified by the model, we conduct the manual annotation to establish the golden standard evaluation corpus. Our approach mitigates the need for extensive annotation efforts by tasking annotators with a binary classification task: distinguishing between automatically corrected sentences and those that remain incorrect. Furthermore, we deeply explore the feasibility of leveraging existing {LLMs}, such as {GPT-3.5-Turbo} and GPT-4, to reduce the annotating workload in grammatical correction tasks. Importantly, our framework is applicable to other languages because the process of constructing the corpus to be annotated does not rely on any language features or language resources. The findings reveal significant potential for enhancing the performance of {LLMs} for low-resource languages.

The main contributions of this paper are:

1. We propose a framework for constructing {the} GEC evaluation corpus.

2. We construct a corpus of Indonesian GEC {tasks}, which extends the existing dataset in the field of Indonesian grammatical correction.

3. We explore the feasibility of existing large-scale models in reducing annotating workload for GEC tasks.

\section{Related Work}

\subsection{Low-resource GEC}

In the realm of GEC tasks, the frequent limitation of low resources has prompted the development of three primary approaches. These approaches concentrated on {the} data-augmentation-based GEC method, unsupervised GEC method based on language models, and few-shot GEC method.

\textbf{Data-augmentation-based GEC method.} For Arabic GEC, \citet{10.1016/j.jksuci.2023.101572} recently introduced a set of seven augmentation techniques: {misspellings}, swap, reverse, replace, mono, token, and source. These methods are distinct from conventional approaches in that their target sentences are generated from real examples, and related enhanced examples are merged together. Compared to the most common data augmentation and synthetic data approaches, the confusion of these {augmentations} outperformed these {methods}. Besides, \citet{app122010380} proposed a semi-supervised method to generate synthetic data. Specifically, the method used three techniques: introducing misspellings, creating punctuation errors via POS tagging, and generating semantic errors by swapping word pairs. These synthesized sentences were then paired with the original ones to create diverse Indonesian GEC training sets.

\textbf{Unsupervised GEC method.} \citet{10.1162/coli_a_00478} conducted a comprehensive survey of GEC tasks and discussed the use of language model-based GEC as a method for low-resource and unsupervised GEC. This approach didn't require training a system with parallel data, but instead used various techniques that used n-gram or Transformer language models.

Transformer-based language models were frequently employed both as Discriminators and Generators. As Discriminators, these language models were assigned the task of transforming low-probability sentences into high-probability ones, effectively reducing errors. This process often involved the utilization of the "break-it-fix-it" (BIFI) method, as proposed by \citet{yasunaga2021break}. BIFI worked by training a fixer and did not require labeled data. By introducing a human-based or learned critic, it could be applied to unsupervised GEC {tasks}. On the other hand, as Generators, these models were primarily employed to generate corrections when faced with limited or zero-sample data. {The} model could correct the errors through the given prompt and noisy input sentence. What's more, \citet{lin2023bertbased} proposed a BERT-based unsupervised GEC framework. The framework treated Tagalog GEC as a multi-class classification task. Instead of using labeled data, the framework evaluated {sentences} using a pseudo-perplexity scoring mechanism. If potential errors were detected based on this score, the system would correct them. Essentially, it used sentence naturalness to identify and fix errors without explicit error labels.

\textbf{Few-shot GEC method.} Nowadays, the application of the few-shot approach in GEC tasks {is} still relatively limited. \citet{zhang2021few} applied meta-learning techniques to the few-shot GEC domain. Remarkably, they completely gave up the use of pseudo {datasets}. Their primary approach involved leveraging the source domain to learn parameters, utilizing the first language (L1) of second language learners in the GEC context. This entailed employing nine L1s (Korean, Traditional Chinese, Japanese, Singapore English, Malay, Burmese, Thai, Vietnamese, and English) as the source domain and five L1s (German, Russian, French, Indonesian, and Mongolian) as the target domain.

\subsection{Research on the Construction of GEC Corpus}

\citet{napoles-etal-2017-jfleg} have significantly contributed to the field by constructing a parallel corpus known as the JHUFLuency-Extended GUG corpus \cite{heilman-etal-2014-predicting}. Firstly, they introduced 1511 GUG sentences (cross-sectional non-grammatical sentences), which were composed by language learners at different proficiency levels and from various L1 backgrounds. Secondly, they collected four manual corrections for each GUG sentence, which in turn formed the JFLEG corpus. Compared with other corpora, the corpus could still be corrected into fluent sentences when the sentence grammar was relatively perfect. 

While focusing on constructing {a} low-resource GEC corpus, \citet{koyama-etal-2020-construction} argued that the Lang-8 corpus was unsuitable as an evaluation dataset due to its inclusion of inappropriate sentences in the corrected text. Consequently, \citet{koyama-etal-2020-construction} developed and published an evaluation corpus specifically aimed at correcting grammatical errors made by Japanese as a Second Language (JSL) learners. Due to its reduced noise and alignment with the dataset's characteristics, the corpus was exceptionally suitable for evaluating and correcting grammatical errors in sentences written by JSL learners. Through manual evaluation, \citet{suzuki2022construction} developed a quality estimation dataset to construct an automatic evaluation model for Japanese GEC. Four GEC systems were utilized to generate corrected sentences for 2,042 sentences written by Japanese learners from the TEC-JL dataset \cite{koyama-etal-2020-construction}. Additionally, a meta-evaluation was performed to assess the dataset's utility in constructing the quality estimation model for the Japanese language. \citet{lin2023bertbased} introduced an Indonesian GEC corpus and a Tagalog GEC corpus. These were the first GEC annotated corpus for Indonesian and Tagalog. It is worth noting that they treated the GEC task as a multi-classification problem, which imposed certain limitations and rendered it unsuitable for evaluating generative GEC models. Therefore, how to build a corpus that faithfully represents the distribution of real-world data while conserving valuable labeling resources has become the core of the GEC task. 

Therefore, in this paper, we focus on how to construct a GEC evaluation corpus that conforms to the real distribution while minimizing the labor-intensive efforts involved.

\subsection{Application of ChatGPT}

The development of generative pre-trained Transformers (GPT) has {shown} remarkable capabilities across various NLP tasks. Consequently, numerous GPT-based approaches have been explored for GEC tasks. For instance, \citet{loem-etal-2023-exploring} primarily employed a prompt-learning-based approach using GPT-3 in GEC tasks. While the advent of GPT-3 has positive impacted on the GEC task, the results obtained still {have} certain limitations. These limitations can be attributed to the fact that the experiment was carried out only using GPT-3.

The ChatGPT, a large-scale language model based on the advanced GPT-3.5 architecture, has demonstrated outstanding potential in diverse NLP tasks. In order to demonstrate the capabilities of ChatGPT in GEC, \citet{fang2023chatgpt} devised zero-shot and few-shot chain-of-thought (CoT) settings by employing in-context learning. Experimental results and human evaluations confirmed the excellent error detection capabilities of ChatGPT, which enabled it to effortlessly correct English errors and enhance the fluency of the corrected sentences. \citet{kwon2023chatgpt} extended their investigation by applying GPT-4 to the GEC task for the low-resource language Arabic. Their work could be considered as a notable improvement, which underscored the remarkable proficiency of LLMs in low-resource language GEC tasks. To augment the input corpus, \citet{fan2023grammargpt} constructed a hybrid dataset containing six distinct error types, augmenting the efficacy of GPT in GEC tasks through the incorporation of a substantial number of pseudo-labels. 

In alignment with these developments, we explore the feasibility of leveraging existing LLMs, such as {GPT-3.5-Turbo} and GPT-4, to alleviate the workload of corpus annotating in GEC tasks.

\begin{figure*}[!ht]
\begin{center}
\includegraphics[scale=0.6]{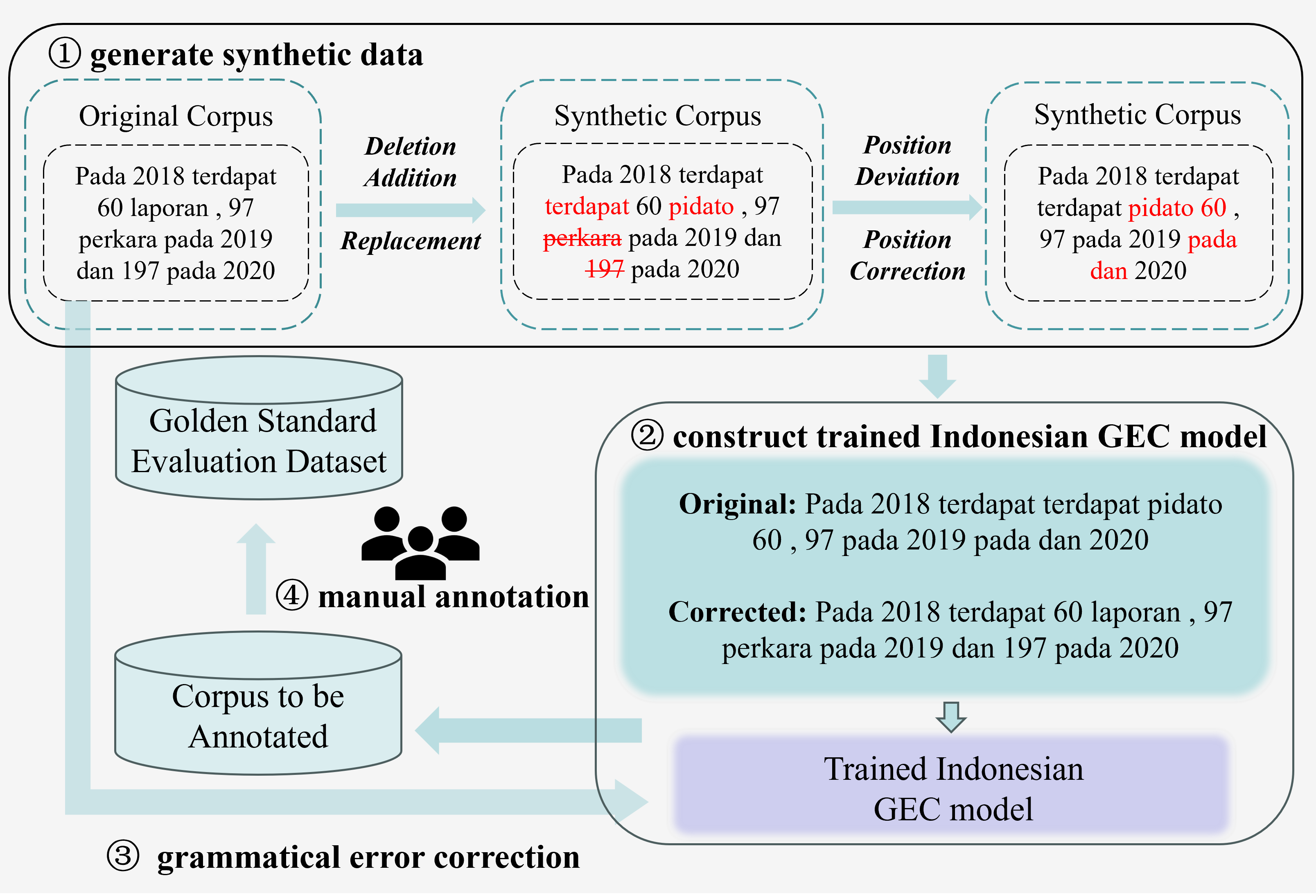}
\caption{An illustration of GEC corpora construction framework.}
\label{fig.1}
\end{center}
\end{figure*}

\section{GEC corpus construction framework}
As illustrated in Figure \ref{fig.1}, we propose a GEC corpus construction framework. In the initial stages of constructing the corpus, we crawl authentic news corpus as the foundation for generating synthetic data with errors. Subsequently, the framework proceeds to train an Indonesian GEC model using the synthetic data. The trained Indonesian GEC model is then employed to correct errors in the authentic news corpus. For the sentences modified by the model, we conduct the manual annotation to establish the golden standard evaluation corpus.

\subsection{Construct of Original Corpus}
\begin{figure*}[!ht]
\begin{center}
\includegraphics[scale=0.35]{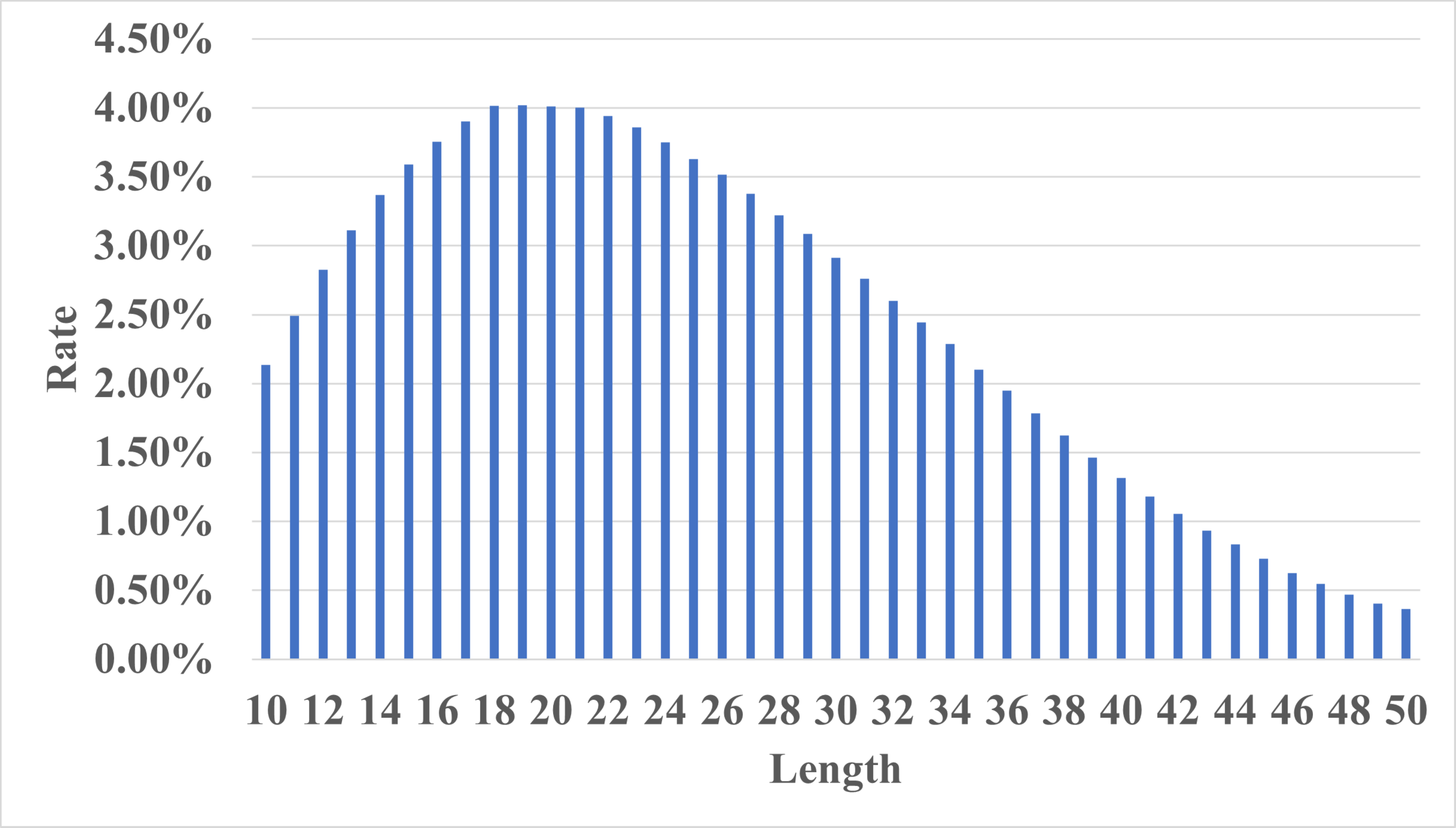} 
\caption{{Length distribution of our selected news corpus.}}
\label{fig.2}
\end{center}
\end{figure*}

We obtain news text from the Antara News Agency\footnote{https://antaranews.com/}, which is Indonesia's sole national news agency. {The corpus is compiled by crawling 631,964 articles from Antara News spanning from 2007 to 2021, across six diverse themes: technology, politics, law, economy, sports, and humanities. Our corpus not only broadens the temporal and topical range of the data but also enhances the granularity and utility of the corpus for robust grammatical error correction.} Subsequently, we employ the Natural Language Toolkit (NLTK)\footnote{https://www.nltk.org/} to realize sentence tokenization and word segmentation on the acquired text. Also, we keep sentences with a length ranging from 10 to 50 words, to ensure that the sentence length is within an appropriate range, thereby effectively improving the learning efficiency of the model. This process generates a total of 7,613,950 sentences as the foundation for constructing the original corpus. Furthermore, we conduct an analysis of the sentence length distribution, the results of which are presented in Figure \ref{fig.2}. Notably, sentence lengths predominantly cluster within the range of 18 to 22 words. {To avoid overfitting, we select 100,000 sentence pairs for the validation set and an additional 100,000 sentence pairs for the test set.}

\begin{table}[h]
\begin{center}
\caption{{The correspondence between common error types and the four operations.}}
\label{tab1}
\begin{tabular}{ccccc}
\hline
{Error Type} & {Deletion} & {Addition} & {Replacement} & {Position Deviation} \\
\hline
% {Verb Tense Error} & {\ding{52}} & {\ding{52}} & {\ding{52}} & {\ding{52}} \\
{Subject Verb Agreement Error} & {\ding{52}} & {\ding{52}} & {\ding{52}} & {\ding{52}} \\
{Word Order Error} &  &  &  & {\ding{52}} \\
{Noun Adjective Agreement Error} & {\ding{52}} & {\ding{52}} & {\ding{52}} & {\ding{52}} \\
{Preposition Error} & {\ding{52}} & {\ding{52}} & {\ding{52}} & {\ding{52}} \\
{Pronoun Error} & {\ding{52}} & {\ding{52}} & {\ding{52}} & {\ding{52}} \\
{Punctuation Error} & {\ding{52}} & {\ding{52}} & {\ding{52}} & {\ding{52}} \\
{Article Error} & {\ding{52}} & {\ding{52}} & {\ding{52}} & {\ding{52}} \\
{Conjunction Error} & {\ding{52}} & {\ding{52}} & {\ding{52}} & {\ding{52}} \\
{Redundancy} &  & {\ding{52}} &  & \\
{Sentence Fragmentation} & {\ding{52}} &  &  & \\
\hline
\end{tabular}
\end{center}
\end{table}

\subsection{Generation of Synthetic Corpus}

We adopt the synthetic data generation method proposed by \citet{DBLP:conf/naacl/ZhaoWSJL19}, with a primary focus on the original corpus to generate synthetic sentences. In this process, we aim to synthesize erroneous sentences, denoted as $sen_{s}=\{x_1^{'},x_2^{'},x_3^{'},…,x_m^{'}\}$, from the original sentences, denoted as $sen_t=\{x_1,x_2,x_3,…,x_n\}$, where $m$ and $n$
represent the lengths of the original sentence and synthetic sentence, respectively. For each token, $x_i$, within the original sentence, we introduce various operations to simulate errors:

a. Deletion: Each token, $x_i$, has a 10\% probability of being deleted, simulating the absence of information.

b. Addition: Similarly, each token has a 10\% probability of being added to simulate the introduction of additional information.

c. Replacement: Each token has a 10\% probability of being replaced by another randomly selected token, introducing lexical variation.

d. Position Deviation: We introduce position deviation using a normal distribution to disrupt the word sequence. Simultaneously, we carry out position correction with a standard deviation of 0.5, adding diversity to word reordering.

{We comprehensively outline the associations between prevalent error types and the four fundamental operations utilized in our analysis, which are shown in Table \ref{tab1}.} By utilizing these operations extensively on {the} original corpus, our GEC model learns to reconstruct input sentences, even when not fully trusting all input information. The sentence pairs generated by this synthesis process have a certain similarity to the sentence pairs encountered in the GEC task {as} both processes involve transforming imperfect sentences into perfect ones by deleting, adding, replacing, or rearranging tokens. {Regarding the non-existent words resulting from character-level morphological modifications, we classify these as spelling errors. Consequently, we do not include this aspect in our framework.}

\subsection{Construction of Indonesian GEC Model}

{The three datasets resulting from the synthesis operation are defined as follows: the training set ($C_t$), the validation set ($C_v$), and the test set ($C_e$).} Based on the constructed synthetic corpus, we develop a Transformer-based Indonesian GEC model, denoted as $M$. The Transformer employs a self-attention mechanism, enabling the model to process input data without relying on sequence order. This approach allows the model to simultaneously process different segments of the input sequence, which {enhances} parallelism and modeling capabilities, consequently, making it excellent when handling lengthy texts. 

We train the model on an RTX TITAN 24G GPU. We implement the GEC model based on Fairseq\footnote{https://github.com/pytorch/fairseq} and PyTorch\footnote{https://github.com/pytorch/pytorch} frameworks. The model runs 50 training epochs with a batch size of 64 and a maximum token limit of 4096 tokens per batch. We employ a triangular learning rate scheduler with a maximum learning rate of 0.004. Gradient clipping with a threshold of 2 is applied to prevent gradient explosions. The encoder and decoder both have 512 word embedding dimension and are both made up of 6 blocks with 8 attention heads in each block and a fully connected layer of 4096 neurons after that. Both the encoder and decoder have a maximum input length of 1024. To mitigate overfitting, we incorporate a 0.2 dropout rate, as well as ReLU dropout and attention dropout, into the model. Additionally, both the encoder and decoder use the shared embeddings. We verify the validation loss for each epoch. The beam search size in the decode phase is set as 5, and the model with the highest $F_{0.5}$ score on the development set is the optimal model.

Following the training on {the} synthetic corpus, we utilize the trained model $M$ to predict corrections for the original corpus. Subsequently, sentences identified by the model as potentially containing errors are extracted from the original corpus, forming a corpus to be annotated $C_a$. We consider that these possible errors are real errors produced by editors in text editing in {the} real world.

\subsection{Manual Annotation}
Since the model obtains a large number of erroneous samples when making predictions on the corpus, it becomes impractical to manually annotate all sentences. Consequently, in the creation of $C_a$, we employ a selection process aiming at identifying errors most likely to occur in real-world scenarios from the corpus to be labeled.

% Based on the three different operations of adding, deleting, and replacing, we select samples whose number of modification operations between the original sentence and the revised sentence is less than or equal to 3.
{In this study, grammatical errors are categorized into nine types based on the nature of the editing operation required to correct them. The categorization includes three subtypes each for missing, misadding, and misusing errors, based on the number of modifications (one, two, or three), ensuring a comprehensive but manageable framework for analysis. The decision to focus on errors with an edit distance of less than three is made to balance the complexity of error correction with the practical capabilities of current grammatical error correction systems.} These samples correspond to nine distinct grammatical error categories as follows: missing one word (missing\_1), missing two words (missing\_2), missing three words (missing\_3), misadding one word (misadding\_1), misadding two words (misadding\_2), misadding three words (misadding\_3), misusing one word (misusing\_1), misusing two words (misusing\_2), and misusing three words (misusing\_3). For sentences with a higher degree of errors, particularly those with a number of {modifications} greater than three, their occurrence frequency is notably lower. Consequently, our attention is primarily directed toward the nine common grammatical error categories. In our selection, we randomly select 2700 samples for manual annotation, specifically 300 samples for each of the specified grammatical error correction types. {To guarantee the accuracy and reliability of the corpus annotations, especially given the complex nature of grammatical error correction in non-native language datasets, we incorporate the expertise of Indonesian language specialists into our evaluation process. These experts, all possessing extensive experience in teaching Indonesian to non-native speakers, are tasked with annotating the labels of two subtasks. This involvement ensures that our dataset evaluation is linguistically sound and reliable.}

Traditional GEC corpus annotation tasks necessitate annotators to revise sentences, ensuring that the modified sentence remains free of errors. This task imposes rigorous requirements on annotators. In this situation, we present a simplified grammatical error correction annotation task. The task of the annotator is to evaluate two specific subtasks. For subtask 1, they should determine whether the correction performed by the model on the sentences we selected is accurate. For subtask 2, they {need to} assess whether the corrected sentence still contains errors.

By simplifying the original GEC annotation task into two distinct discrimination tasks, the overall complexity of the annotation task thus can be effectively reduced. subtask 1 evaluates the accuracy of the model's corrections on selected sentences, while subtask 2 examines the presence of residual errors in the corrected sentences. A correct correction operation in subtask 1 does not ensure the complete absence of errors, so a sentence may still contain mistakes after a partially correct correction. The two subtasks are indeed related, but they aim to address different aspects of the grammatical error correction process. Consequently, it is necessary to further filter the samples to identify sentences deemed entirely correct in subtask 2.

\subsection{{Consistency between Our Corpus and Real-world Data}}

{In addressing the challenge of bridging the gap between dataset errors and those found in real-world applications, our approach integrates both synthetic and real-world data processing. Initially, our model is trained extensively on synthetic datasets designed to simulate a wide range of grammatical errors. Following this, the model is tasked with identifying and correcting errors in a corpus of real-world news articles, which is continuously updated to reflect current language usage trends. Each potential error identified by the model, along with the suggested correction, is subjected to a rigorous review process by linguistic experts. This dual-validation mechanism ensures that our dataset not only reflects real-world grammatical inaccuracies but also that the corrections are contextually appropriate and linguistically accurate. This methodology ensures our dataset remains a reliable resource for developing robust grammatical error correction systems.}

\section{Self-constructed Corpus}

\begin{table}[h]
\begin{center}
\caption{Distribution of Indonesian GEC Corpus.}
\label{tab2}
\begin{tabular}{cc}
\hline
Type & Number \\
\hline
misadding one word & 245 \\
missing one word & 288 \\
misusing one word & 254 \\
misadding two words  & 264 \\
missing two words & 292 \\
misusing two words & 255 \\
misadding three words & 244 \\
missing three words & 275 \\
misusing three words & 247 \\
\hline
Sum & 2364 \\
\hline
\end{tabular}
\end{center}
\end{table}

% \begin{figure}[!ht]
% \begin{center}
% \includegraphics[scale=0.5]{fig44.png} 
% \caption{Annotation results of subtask 1. Since there are some sentence pairs where it is impossible to determine whether the correction operation is true, the total number of samples in some categories is less than 300.}
% \label{fig.4}
% \end{center}
% \end{figure}

% \begin{figure}[!ht]
% \begin{center}
% \includegraphics[scale=0.5]{fig55.png} 
% \caption{Annotation results of subtask 2. Since there are some sentence pairs where it is impossible to determine whether the corrected sentence still contains errors, the total number of samples in some categories is less than 300.}
% \label{fig.5}
% \end{center}
% \end{figure}

\begin{figure}[htbp]  
\begin{center}  
\subfigure[{Annotation results of subtask 1. Since there are some sentence pairs where it is impossible to determine whether the correction operation is true, the total number of samples in some categories is less than 300.}]{  
\includegraphics[width=0.48\linewidth]{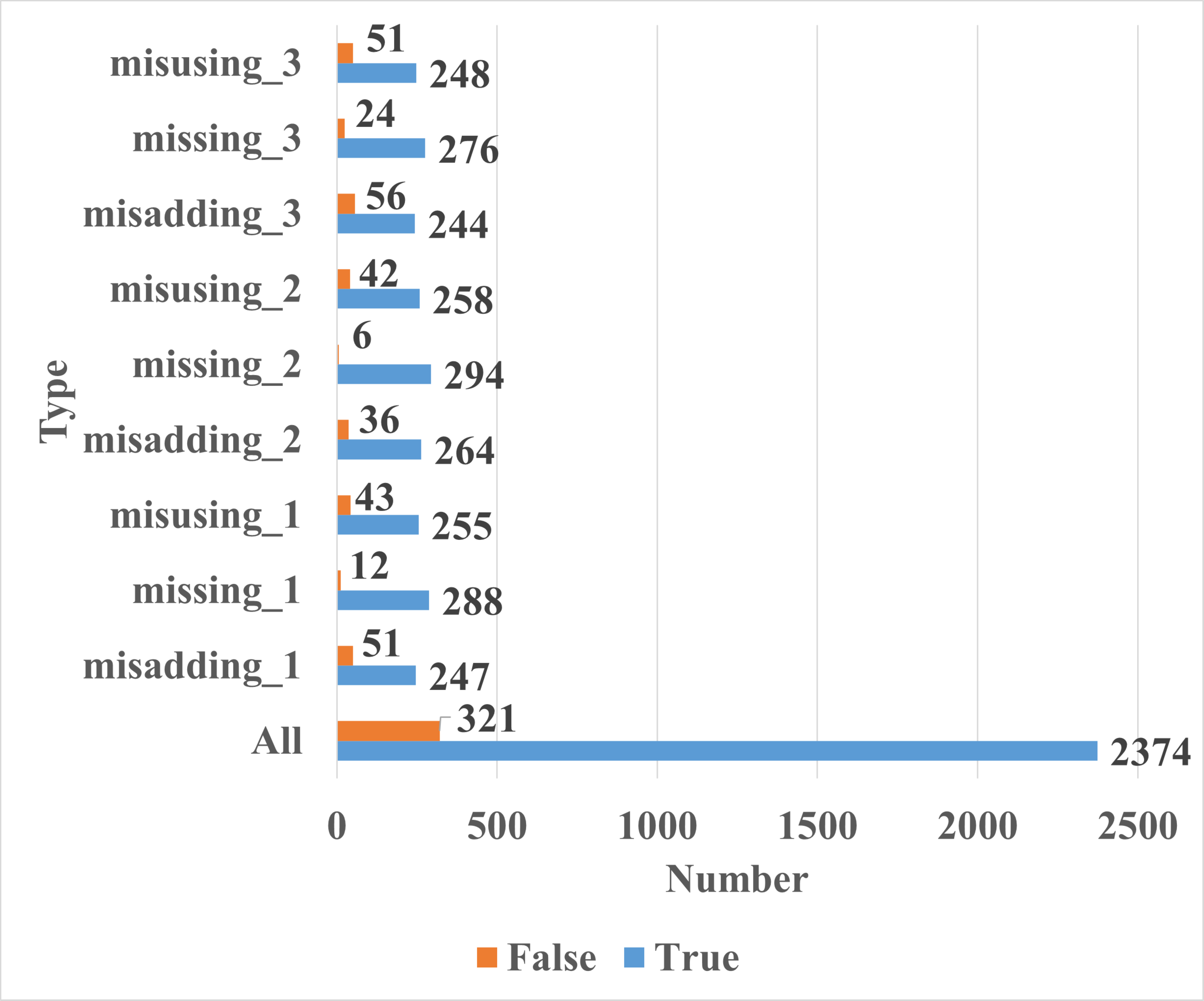}  }  
\subfigure[{Annotation results of subtask 2. Since there are some sentence pairs where it is impossible to determine whether the corrected sentence still contains errors, the total number of samples in some categories is less than 300.}]{  
\includegraphics[width=0.48\linewidth]{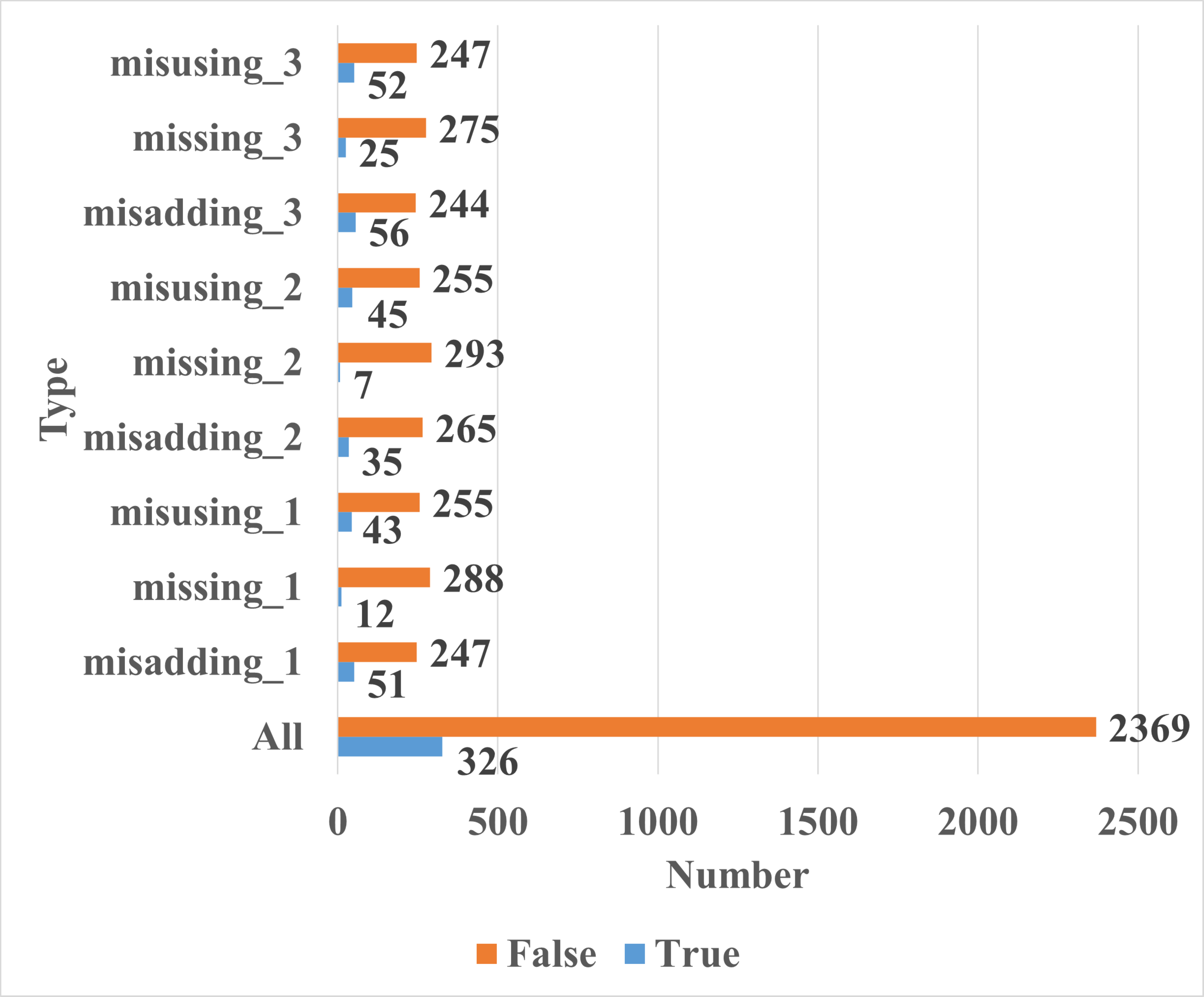}  }  
\caption{{Annotation results of each subtask.}}  
\label{fig.3}
\end{center} 
\end{figure}

After manual annotation, we only retain samples where the correction operation is correct and the corrected sentences have no errors. These samples constitute our Indonesian GEC corpus. The data statistics of the corpus are shown in Table \ref{tab2}. There are 2,364 samples that contain nine different error types in total, representing 87.6\%, contained grammatical errors in the source sentences and were subsequently corrected. This indicates that the majority of the samples provided to the annotators are suitable for constructing a GEC corpus, obviating the need for annotators to sift through numerous error-free samples to identify and accurately correct those with grammatical errors. Obviously, the error type of deleting holds the biggest proportion, with 288 missing one word (missing\_1), 292 missing two words (missing\_2), and 275 missing three words (missing\_3), respectively. 

In this task, annotators are tasked with assessing two specific aspects: first, they determine whether the correction performed by the model on the sentence is accurate (subtask 1), and second, they assess whether the corrected sentence still contains errors (subtask 2). 

The results of subtask 1 are presented in Figure \ref{fig.3} (a), encompassing a total of 2,695 samples. Among these samples, 2,374 exhibited correct modifications, while 321 displayed incorrect modifications. Obviously, the accuracy of modifications was relatively high. Within this dataset, when considering the total number of samples for each error type (which stands at 300 samples per type), the highest accuracy in error modification is observed for deletions. In this category, the average number of accurately modified samples {surpasses} that of the other two types. Particularly it is more accurate in modifying sentences where two words are missed, with an impressive count of 294 samples. This may be due to the fact that the feature of deletion is more obvious and easy to capture and modify. 
% In general, the error correction effect of this GEC framework is relatively good.

In addition, the results of subtask 2 are presented in Figure \ref{fig.3} (b). {Of} 2695 samples, there are 2369 samples without any errors and 326 samples still contain errors. Among these samples with errors, the misadding errors are relatively more than other errors, specifically, when three words are misadded, with {a} count of 56. However, there are only 7 samples {that} contained missing errors when two words are missed, demonstrating the GEC framework is quite effective in {correcting} missing errors and it is consistent {with} the result of subtask 1. 
% In all, only a small amount of samples still contain errors, so this GEC framework can be widely used in GEC work.

The results from the annotations indicate that within the corpus selected by our framework, a significant portion of the sentences in the source language exhibit errors and these errors can be accurately rectified, which {suggests} that it is feasible to construct a GEC corpus while substantially decreasing the workload for annotators.

\section{Performance Verification of Large-scale Language Model}

We investigate the feasibility of utilizing existing LLMs {(GPT-3.5-Turb, GPT-4, PolyLM, LLAMA3, and InternLM2)} to streamline corpus annotation efforts in GEC tasks. {It is noted that our analysis of the GEC datasets diverges from traditional grammatical error correction benchmarks. We focus specifically on two binary classification subtasks using {LLMs}. This evaluation is conducted at the sentence level, employing sentence-level evaluation metrics to gauge the performance of the models in these specific contexts. This methodological focus is intended to explore the feasibility of using {LLMs} for precise tasks within the error correction domain, rather than evaluating their overall performance in correcting grammatical errors across a text.} 

\subsection{Datasets}
We randomly selected 100 samples of each error type from the constructed Indonesian corpus to construct a large-scale language model evaluation corpus. In addition, in order to verify the effectiveness of our prompt template, we construct 100 examples from the test set of the CoNLL14 GEC corpus \cite{ng-etal-2014-conll} as the English evaluation corpus. The label distribution of evaluation corpora is shown in Table \ref{tab3}{\footnote{{In Table \ref{tab3}, the total number of samples listed as 895 reflects a minor adjustment from the initially planned 900 samples. This adjustment was due to the exclusion of 5 samples that our expert annotators identified as ambiguous, thereby preventing a definitive judgment on their grammatical correctness. This measure was taken to ensure the precision and reliability of our dataset and subsequent model evaluation.}}}.

\begin{table}[h]
\begin{center}
\caption{Distribution of Evaluation Corpus.}
\label{tab3}
\begin{tabular}{cccc}
\hline
Task & Label & English & Indonesian \\
\hline
\multirow{2}{*}{subtask 1} & False & 50 & 123 \\
 & True & 50 & 772 \\
 \hline
\multirow{2}{*}{subtask 2} & False & 50 & 766 \\
 & True & 50 & 129 \\
\hline
\end{tabular}
\end{center}
\end{table}

\subsection{Models}

{We evaluate the performance of five large-scale models on two subtasks. The models included two closed-source models, GPT-3.5-Turbo\footnote{https://chat.openai.com} and GPT-4 \cite{openai2023gpt4}, as well as three open-source models: PolyLM \cite{wei2023polylmopensourcepolyglot}, Llama3\footnote{https://github.com/meta-llama/llama3}, and InternLM2 \cite{cai2024internlm2technicalreport}.}

\begin{figure*}[!ht]
\begin{center}
\includegraphics[scale=0.45]{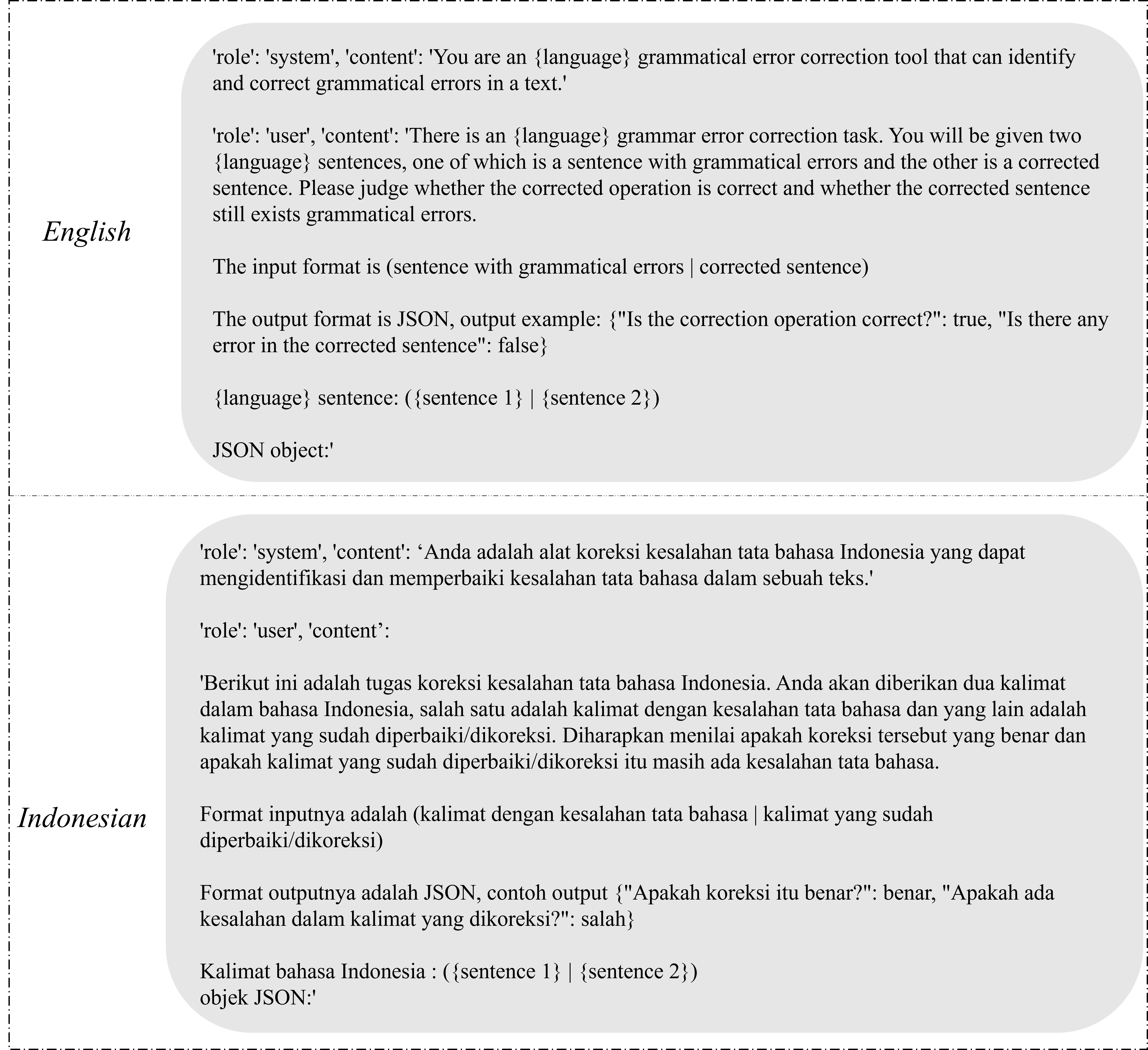} 
\caption{An illustration of the prompt template used for {GPT-3.5-Turbo} and GPT-4. \{language\} is the language of the input sample. \{sentence 1\} and \{sentence 2\} respectively represent the sentence with grammatical errors and the corrected sentence.}
\label{fig.4}
\end{center}
\end{figure*}

\textbf{{GPT-3.5-Turbo}\footnote{https://chat.openai.com}}: {GPT-3.5-Turbo} is a cutting-edge language model created by OpenAI. It leverages the power of artificial intelligence and natural language processing to provide intelligent and contextually relevant responses to text-based queries and prompts. With a vast knowledge base acquired from diverse internet sources, {GPT-3.5-Turbo} can engage in meaningful conversations, answer questions, offer explanations, generate creative content, and assist with a wide range of tasks. 
% We choose {GPT-3.5-Turbo} as the evaluated model, which stands out as the most advanced and specifically optimized for chat functionality.

\textbf{GPT4 \cite{openai2023gpt4}}: GPT-4 is a multimodal model capable of processing both image and text inputs, and generating text outputs. GPT-4 is a pre-trained Transformer-based model specifically designed to predict the subsequent token in a given document. Through the post-training alignment process, GPT-4 achieves enhanced performance in terms of factual accuracy and adherence to desired behavior.

{\textbf{PolyLM \cite{wei2023polylmopensourcepolyglot}} is an advanced open-source language model developed with a focus on versatility and performance. One of PolyLM's standout features is its robust multilingual capabilities. It is designed to understand and generate text in multiple languages, making it a versatile tool for global applications. This multilingual proficiency ensures that PolyLM can be effectively utilized in diverse linguistic environments, further enhancing its utility and reach.}

{\textbf{LLAMA3\footnote{https://github.com/meta-llama/llama3}} is a state-of-the-art open-source language model developed to advance natural language processing capabilities. Engineered with cutting-edge techniques, LLAMA3 excels in a variety of applications, including text generation, comprehension, and translation. LLAMA3's design emphasizes high performance, accuracy, and adaptability, ensuring it meets the complex demands of modern AI-driven language tasks.}

{\textbf{InternLM2 \cite{cai2024internlm2technicalreport}} is an advanced open-source large language model designed to surpass previous models in performance across various benchmarks. It excels in handling long-context data and incorporates advanced alignment methods. InternLM2's development includes meticulous data preparation and diverse training stages, providing the AI community with a powerful tool for natural language processing tasks.}

\begin{figure*}[!ht]
\begin{center}
\includegraphics[scale=0.5]{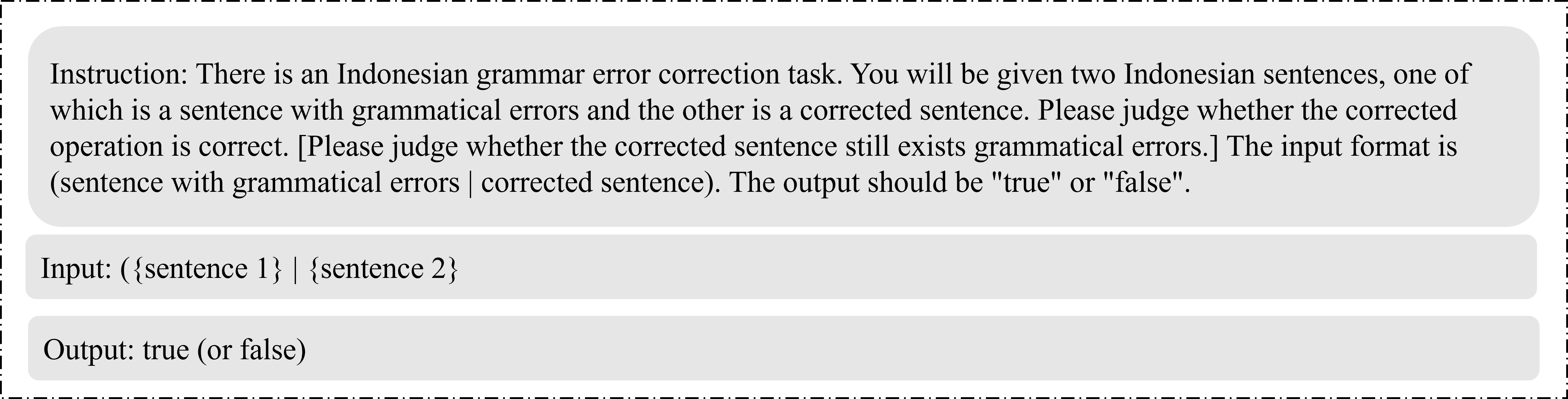} 
\caption{{An illustration of the prompt template used for open-source {LLMs}. \{sentence 1\} and \{sentence 2\} respectively represent the sentence with grammatical errors and the corrected sentence. [] is the sentence used in subtask 2.}}
\label{fig.6}
\end{center}
\end{figure*}

{The specific settings used for GPT-3.5-Turbo and GPT4 during the experiments are as follows. The temperature is set as 1. The presence penalty and frequency penalty are both set to 0, ensuring no artificial boost or suppression of particular words based on their occurrence in the context. For the three open-source {LLMs}, we set the temperature to 1 and the number of beams to 4.} The prompt template of {GPT-3.5-Turbo} and GPT4 is shown in Figure \ref{fig.4}. {In addition, the prompt template of open-source {LLMs} is presented in Figure \ref{fig.6}.}

\begin{table*}[h]
\begin{center}
\caption{Experimental Results of subtask 1.}
\label{tab4}
\begin{tabular}{ccccc}
\hline
Language & Model & P & R & F \\
\hline
\multirow{6}{*}{Indonesian} & {GPT-3.5-Turbo} (English Template) & {0.8680}  & {0.9793}  & {0.9203} \\
 & {GPT-3.5-Turbo} (Indonesian Template) & {0.8636}  & {0.9922}  & {0.9234} \\
 & GPT4 (English Template) & {0.9074}  & {0.8886}  & {0.8979} \\
  & {PolyLM-13b (English Template)} & {0.8429} & {0.0764} & {0.1401} \\
 & {Llama3-8b (English Template)} & {0.8250} & {0.7267} & {0.7727} \\
 & {Internlm2-20b (English Template)} & {0.8609} & {0.9858} & {0.9191} \\
 \hline
\multirow{2}{*}{English} & {GPT-3.5-Turbo} & 0.6912  & 0.9400  & 0.7966 \\
 & GPT4 & 0.7925  & 0.8400  & 0.8155 \\
\hline
\end{tabular}
\end{center}
\end{table*}

\begin{table*}[h]
\begin{center}
\caption{Experimental Results of subtask 2.}
\label{tab5}
\begin{tabular}{ccccc}
\hline
Language & Model & P & R & F \\
\hline
\multirow{6}{*}{Indonesian} & {GPT-3.5-Turbo} (English Template) & {0.3600}  & {0.1395}  & {0.2011} \\
 & {GPT-3.5-Turbo} (Indonesian Template) & {0.2857}  & {0.0155}  & {0.0294} \\
 & GPT4 (English Template) & {0.3879}  & {0.3488}  & {0.3673} \\
 & {PolyLM-13b (English Template)} & {0.0476} & {0.0078} & {0.0133} \\
 & {Llama3-8b (English Template)} & {0.0915} & {0.5891} & {0.1583} \\
 & {Internlm2-20b (English Template)} & {0.1359} & {0.9302} & {0.2372} \\
\hline
\multirow{2}{*}{English} & {GPT-3.5-Turbo} & 0.8864  & 0.7800  & 0.8298 \\
 & GPT4 & 0.8261  & 0.7600  & 0.7917 \\
\hline
\end{tabular}
\end{center}
\end{table*}

\subsection{Results}
The experimental results of subtask 1 and subtask 2 are shown {in} Table \ref{tab4} and Table \ref{tab5}. {The performance of the five large-scale models on two subtasks is analyzed as follows.}

{In subtask 1 for Indonesian, we use the English and Indonesian templates respectively on GPT-3.5-Turbo and only use the English template for the other four {LLMs}. Specifically, the average scores for the three metrics are 0.8613, 0.7748, and 0.7623, respectively. On GPT-3.5-Turbo, using either English or Indonesian templates shows no significant impact on model performance. GPT4 achieves the highest precision of 0.9074, and GPT-3.5-Turbo achieves the highest precision and F1 of 0.9922 and 0.9234, respectively. Within these large language models, PolyLM-13b exhibits notably low recall and F1 scores, 0.0764 and 0.1401 respectively. In subtask 1 for English, precision and F1 scores on GPT4 are better than GPT-3.5-Turbo, but not as good as subtask 1 in Indonesian overall. In all, the results fully demonstrate the effectiveness of the large language model in identifying whether corrections are reasonable.}

{For subtask 2, we use the English and Indonesian templates respectively on GPT-3.5-Turbo for Indonesian, and only use the English template for the other four models. Overall, the effect of subtask 2 is not as good as that of subtask 1. Specifically, the average scores for the three metrics are 0.2181, 0.3384, and 0.1678, respectively. The two different templates show significant differences in GPT-3.5-Turbo, with the English template performing better than the Indonesian template. The reason is that GPT-3.5-Turbo mainly uses English data for training and is more suitable for the processing of English templates. It is worth noting that the recall rate on Internlm2-20b is higher at 0.9302, which means that the model will recognize examples that are actually incorrect as correct. Therefore, annotators need to focus on {a} comprehensive evaluation of examples that are recognized as correct. However, for English subtask 2, GPT-3.5-Turbo has high performance, with precision (0.8864) and recall rate (0.7800), F1 score of 0.8298, and GPT-4 also has such high performance in English, indicating that {GPT-3.5-Turbo} and GPT-4 have better performance in English. Experimental results show that these {LLMs} still need to be improved in their ability to identify Indonesian grammatical errors.}

{On the whole, for subtask 1, in both English and Indonesian, the model has a better effect, while for subtask 2, the model has a worse effect on Indonesian and a better effect on English. However, for different language templates, subtask 1 is not affected by templates, while subtask 2 works better on English templates. This difference can generally be attributed to language characteristics, data availability, and model development, specifically, Indonesian has more complex grammar rules and fewer language resources than English.  Although the relevant experimental results {in} Indonesian show that LLMs have {a} certain effectiveness in completing tasks in Indonesian, there is significant room for improvement. Therefore, the annotation of low-resource languages still requires manual annotation to complete.}

\section{Conclusion}
We introduce a framework for constructing {} GEC corpus. Our research focuses on Indonesian as the target language, and we utilize the framework to create an evaluation corpus specifically for Indonesian GEC. This addresses the limitations of existing datasets available for Indonesian. Additionally, we explore the feasibility of leveraging existing {LLMs} such as {GPT-3.5-Turbo} and GPT-4, to streamline the annotation process in GEC tasks. The findings indicate the promising potential for improving the performance of {LLMs} in low-resource language contexts. In the future, we would expand our corpus to encompass additional languages. {Concurrently, we would explore a range of prompt strategies to determine the most effective methods for enhancing model performance in Indonesian GEC tasks, which include a systematic comparison of different prompting techniques.}

\begin{acks}
This work was supported by the Guangdong Philosophy and Social Science Foundation Regular Project (No. GD20CWY10).
\end{acks}

% the National Social Science Fund of China (No.22BTQ068), and the Science and Technology Program of Guangzhou (No.202002030227).
\bibliographystyle{ACM-Reference-Format}
\bibliography{sample-base}

% \end{CJK*}
\end{document}